\documentclass[runningheads]{llncs}

\usepackage{graphicx}
\graphicspath{{images/}}

\usepackage[numbers]{natbib}
\usepackage{microtype, booktabs, multicol, multirow}
\usepackage[pdftex, pagebackref=true, pdfpagelabels=true,
hypertexnames=true, plainpages=false, naturalnames=false,
bookmarks=true, bookmarksnumbered=true,
pdfpagemode=UseOutlines,]{hyperref}
\usepackage{soul} 

\usepackage{amsmath}
\usepackage{tabularx}
\hypersetup{colorlinks, citecolor=[rgb]{0,0.7,0}, filecolor=[rgb]{0.6,0.0,0.0}, linkcolor=[rgb]{0.0,0.0,0.5}, urlcolor=[rgb]{0.6,0.0,0}}

\begin{document}

\title{DermDiff: Generative Diffusion Model for Mitigating Racial Biases in Dermatology Diagnosis}

\titlerunning{Mitigating Racial Biases in Dermatology Diagnosis}

\author{Nusrat Munia \and
Abdullah-Al-Zubaer Imran}
\authorrunning{N. Munia and A. Imran}

\institute{University of Kentucky, Lexington, KY 40506, USA \\
\email{\{nusrat.munia, aimran\}@uky.edu}
}

\maketitle              
\begin{abstract}

Skin diseases, such as skin cancer, are a significant public health issue, and early diagnosis is crucial for effective treatment. Artificial intelligence (AI) algorithms have the potential to assist in triaging benign vs malignant skin lesions and improve diagnostic accuracy. However, existing AI models for skin disease diagnosis are often developed and tested on limited and biased datasets, leading to poor performance on certain skin tones. To address this problem, we propose a novel generative model, named DermDiff, that can generate diverse and representative dermoscopic image data for skin disease diagnosis. Leveraging text prompting and multimodal image-text learning, DermDiff improves the representation of underrepresented groups (patients, diseases, etc.) in highly imbalanced datasets. Our extensive experimentation showcases the effectiveness of DermDiff in terms of high fidelity and diversity. Furthermore, downstream evaluation suggests the potential of DermDiff in mitigating racial biases for dermatology diagnosis. Our code is available at \url{https://github.com/Munia03/DermDiff}.

\end{abstract}
\section{Introduction}
Skin cancer is the most prevalent cancer in not only the United States but also worldwide. Every day, about 9,500 Americans receive a skin cancer diagnosis and every hour more than two people die from this disease \citep{ACS2024}. Early detection of melanoma has a five-year survival rate of over 99 percent \citep{ACS2024}. With the rapid development of machine learning models, Artificial Intelligence (AI) diagnostic tools have significantly improved the detection of malignancy in skin diseases. These models need to be trained on a large pool of labeled data for optimal performance. While curating large labeled dermatology datasets is challenging, existing methods are often limited to biased predictions. Recent research suggests models are biased toward some subgroup populations with sensitive attributes, such as skin tone, age, gender, sex, etc. \cite{Tschandl2018_HAM10000,kinyanjui2020fairness,groh2021fitzpatrick}. For example, models perform poorly on images of darker skin tones compared to those with lighter skin tones. The primary reason behind this can be attributed to training on imbalanced datasets where images of certain racial groups have minority representations. Moreover, images of common skin cancers of darker skin tones are rarely depicted. 

To ensure fairness in skin disease classification, several methods have been proposed \cite{safe,detecting_melanoma,fairprune,ddi,me_fairprune,fairadabn}. \citet{safe} proposed an ensemble approach and trained two different models for lighter and darker skin tones. \citet{detecting_melanoma} applied a debiasing mechanism by unlearning skin tone features from the images to mitigate skin tone bias. \citet{fairadabn} presented adaptive batch normalization for sensitive attributes with a loss function to minimize the difference in prediction probability between subgroups. Although all these models improve fairness performance to some extent, there is still a performance gap due to the lack of a large diverse dataset. \citet{ddi} published the Diverse Dermatology Images (DDI) dataset but it is not large enough to effectively train a deep learning model. \citet{de_biasing2022} adopted a GAN-based augmentation method for debiasing common artifacts like hair, ruler, frame, etc., but without considering demographic attributes. 

Demographic biases exist primarily due to the lack of properly distributed dermatology datasets. Several publicly available datasets include ISIC \cite{isic2018skin,Tschandl2018_HAM10000,isic2020}, Fitzpatrick17k \cite{groh2021fitzpatrick}, DDI dataset \cite{ddi}, and Derm7pt \cite{atlas2018seven}. These datasets, however, do not contain enough demographic information. Among all, ISIC is the largest data source containing $>50k$ image samples with their skin conditions. Unfortunately, no skin type or race-related information is available in the metadata. \citet{ita_skin_tone} proposed to predict the skin tone based on the Individual Typology Angle (ITA) score of the ISIC 2018 \cite{isic2018skin} dataset and shows a very skewed distribution over skin tone. However, experimenting with the human-annotated Fitzpatrick dataset revealed that 
 the ITA-based skin tone detection is not reliable \citep{groh2021fitzpatrick}. 

\begin{figure}[t]
    \centering
    \includegraphics[width=\linewidth]{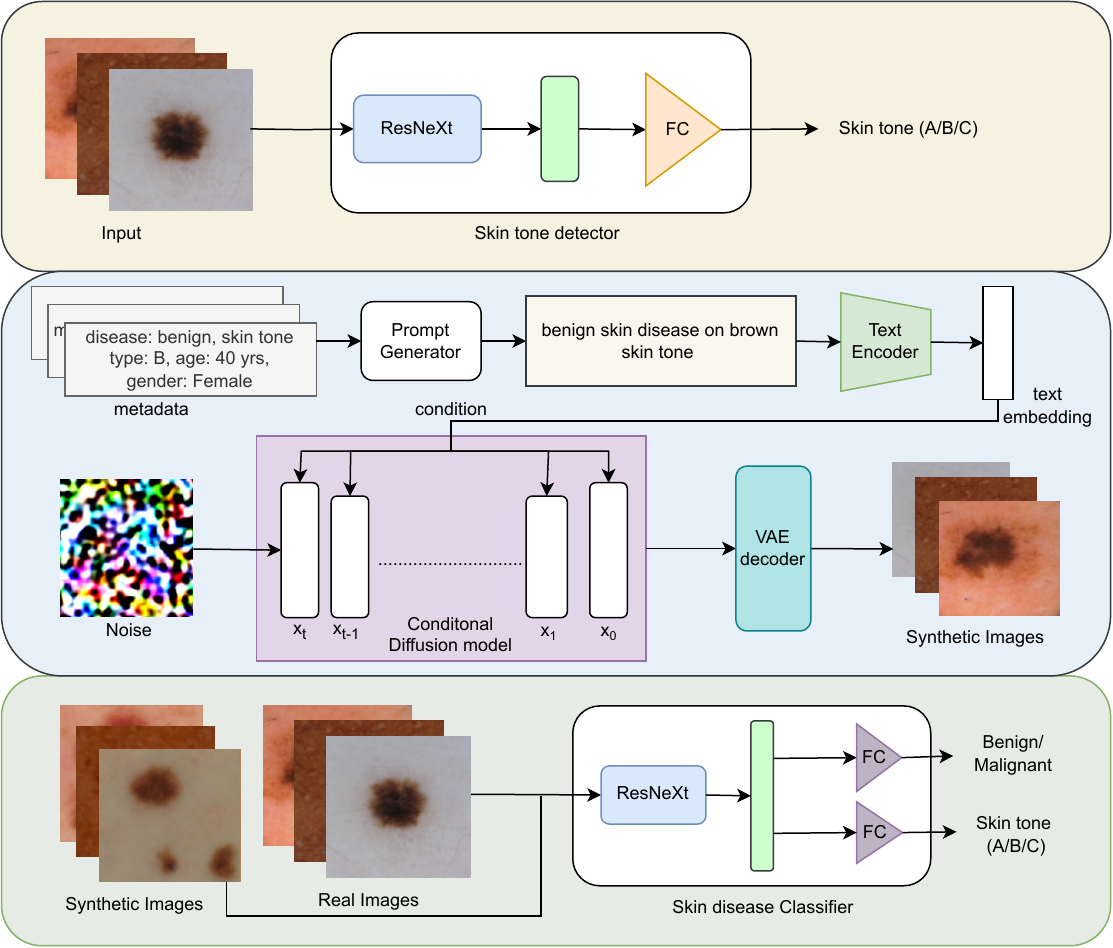}
    \caption{Proposed DermDiff framework: skin tone detector identifies the patient races based on dermoscopic images in a dataset; race and other attributes are used to condition on image generation in the Diffusion-based generative model; and finally Skin disease diagnosis is performed on both real and generated synthetic images.}
    \label{fig:arch1}
\end{figure}

Recently, generative models such as Generative Adversarial Networks (GAN) and Latent diffusion models (LDM) have become popular for generating new sample images similar to real datasets \citep{imran2021multi, rombach2022high}. LDM has shown better performance gains for generating high-quality images \cite{stable_diff}. The denoising process of diffusion models in the latent space proved to have the ability to generate high-resolution, high-fidelity, and diverse images. They can also be trained with conditioning mechanisms that allow control over the inference process for generating new samples. This provides a potential solution for augmentation in a supervised learning pipeline where the training dataset is limited. 

In this work, we leverage the power of generative models to create a diverse dermatology dataset for diagnostic model training. Our proposed DermDiff is conditioned on different image attributes, such as skin tone and disease type. Our specific contributions can be summarized as below:
\begin{itemize}
    \item A generative model framework DermDiff, conditioned on text prompts combining different patient and disease attributes. 
    
    \item A skin tone detection model to categorize dermoscopic images into lighter, brown, or darker tones.
    
    \item Effective diagnosis of skin disease and detection of skin tones in a multi-class classification scheme. 

\end{itemize}

\section{Method}

\subsection{DermDiff}
We propose a Diffusion-based DermDiff model for generating a large number of dermoscopic image samples of diverse attribute sets (Fig.~\ref{fig:arch1}). We train our DermDiff model conditioning on text prompts. First, we collect all the available datasets and their metadata. From available demographic attributes, we generate a text prompt for each image. We only consider skin tone information and disease type in our proposed model. A text encoder is used to map the input text prompt into embedding space with different attribute feature representations. We utilize the Contrastive Language-Image Pre-Training (CLIP) \cite{clip_text} method for text encoding. This embedding of text prompt is used as the condition for training the diffusion model. During training, the dermoscopic images are transformed and mapped to latent representation using a variational autoencoder (VAE) \citep{vae_enc}. 

For a pair of image $x$ and text prompt $y$, a random Gaussian noise $\epsilon$ is sampled from the latent space $\epsilon\sim\mathcal{N}(0, I)$. The text prompt $y$ is encoded using the CLIP text encoder and the image $x$ is transformed into the latent dimension using VAE. For a denoising step $t$, the U-Net-based diffusion model tries to predict the sample noise $\tilde{\epsilon}$ from the encoded text and image representations with added noise. 
\begin{equation}
    \tilde{\epsilon} = {UNet}(VAE(x)\oplus\epsilon, t, CLIP(y)).
\end{equation}
The model is trained with mean squared error loss between sampled noise and predicted noise.
\begin{equation}
    \mathcal{L}_{mse} = \frac{1}{d} \sum_{i=0}^{d-1} (\tilde{\epsilon_{i}} - \epsilon_{i})^2,
\end{equation}
where, $d$ denotes the dimensionality of sampled noise in latent space. We experiment with the pre-trained version of the Stable diffusion model \cite{stable_diff}. The pre-trained text-to-image model (version 1.4 of 1.45B parameters) was trained with LAION-400M dataset \citep{laion}. We fine-tune the pre-trained text-to-image model on the dermatology dataset. 

\subsection{Skin Tone Detection}
We build and train a skin tone detection model. We redefine simplifying the Fitzpatrick skin types (FST) \citep{fitzpatrick1988validity} categories. FSTs include 6 categories of skin tones: 1 to 6 where 1 means the lightest skin tone and 6 means the darkest skin tone. We define three skin tone labels: A--lighter skin tone (FST I-II), B-brown skin tone (FST III-IV), and C-darker skin tone (FST V-VI). We leverage the ResNexT-101 model \citep{resnext} for feature extraction and add an FC layer for three skin tone classification. A focal loss is used to train the classification model. 
\begin{equation}
\mathcal{L}_{focal} = -\alpha (1 - \hat{p})^\gamma \log(\hat{p}),
\end{equation}
where, $\hat{p}$ is the model predicted probability, $\alpha$ and $\gamma$ are the hyperparameters.
\subsection{Dermatology Disease Classification}
After generating a large number of synthetic images, we create a balanced and well-distributed dataset combining both real and synthetic data. We build a multi-class classifier to predict both skin tone and disease status. The classification model with convolutional architecture, ResNeXt-101 \cite{resnext} combined with a classification layer. The model predicts if the sample image is benign or malignant.

\section{Experimental Evaluation}
\subsection{Data}
Table~\ref{tab:data-dist} overviews the datasets used in our experiments. \textbf{Fitzpatrick17k:} This dataset \citep{groh2021fitzpatrick} contains 16,577 skin condition images with Fitzpatrick skin types (FST) \citep{fitzpatrick1988validity} annotation. Among all the images only 4,311 images are neoplastic or tumorous and the rest of them are non-neoplastic images. Use the whole Fitzpatrick dataset for the skin tone classification model. However, for DermDiff and disease classification, we use only neoplastic images. \textbf{ISIC:} The International Skin Imaging Collaboration (ISIC) datasets [2016-2020] \citep{isic2018skin, Tschandl2018_HAM10000, isic2020} are the most popular dataset for skin disease diagnosis. They comprise thousands of dermoscopic images, in addition to gold-standard disease diagnostic metadata. Although their metadata do not include any skin tone information they contain other attributes such as sex, approximate age, etc. \textbf{DDI:} The Diverse Dermatology Images (DDI) dataset \citep{ddi} is a more diverse dataset compared to the former two datasets. It contains a total of 656 dermatology images of three different skin tone categories FST I-II, FST III-IV, and FST V-VI. \textbf{Derm7pt}: The 7-point criterion dataset, also known as atlas dataset \citep{atlas2018seven} contains 1,011 dermoscopic images, and 1,011 clinical images each with metadata and a 7-point checklist criterion. \textbf{MClass}: The MClass \citep{mclass_data} dataset provides a classification benchmark with a total of 200 samples annotated by 157 dermatologists on the images in the dataset. It contains 100 dermoscopic images and 100 clinical images. 

\begin{table}[t]
\centering
\caption{Distribution of dermatology datasets used in our experiments.}
\label{tab:data-dist}
\setlength{\tabcolsep}{4pt}
\begin{tabular}{@{}lc ccccc@{}}
\toprule
Dataset & \phantom{a} & Size & Benign & Malignant & \#Skin tones & Mode \\ 
\midrule
Fitzpatrick && 16,577 & 2,234  & 2,263 & 6 & Train \\
ISIC  && 57,964 & 52,874 & 5,090 & --            & Train \\
DDI && 656 & 485 & 171 & 3 & Test  \\
Derm-7pt && 2,022 & 1,518 & 504 & -- & Test  \\
MClass && 200 & 160 & 40 & -- & Test  \\ 
\bottomrule
\end{tabular}
\end{table}

\subsection{Implementation Details}
\textbf{Inputs:} The input dermoscopic images are 0--1 normalized and resized to $256\times256$ resolution.  
\textbf{Training:} We trained DermDiff using the Fitzpatrick dataset. We also trained the skin tone classification model with the Fitzpatrick datasets and tested it on the DDI dataset. With our skin tone classification model, we predicted skin tones for test datasets. Our models were implemented in Python with the Pytorch library and executed on an Intel(R) Xeon(R) 128GB machine with two NVIDIA RTX A4000 GPUs.
\textbf{Hyper-parameters:} We trained our generative model with a mini-batch size of 4 and a learning rate of $1e^{-4}$ for 10k steps. For both skin tone and melanoma classification we have applied the Focal loss function and Stochastic Gradient Descent (SGD) optimizer with momentum and learning rate of $3e^{-4}$. For skin tone detection, we set the focal loss hyper-parameter $\alpha = [0.3, 0.4, 0.9]$ and $\gamma = 2$. And for disease classification, $\alpha$ is set to 0.8 and $\gamma$ is set to 2. The classification models were trained with a mini-batch size of 64 for 30 epochs with the Fitzpatrick dataset and 4 epochs with other datasets. 
\textbf{Evaluation:} For evaluating the image generation performance, we calculated Fr\'echet Inception Distance (FID) and multi-scale structural similarity index metric (MS-SSIM) scores. FID measures the fidelity of the model, which depicts how close the distribution of generated and real samples are. MS-SSIM ranging between 0 and 1, measures how diverse the generated images are. For classification models, we report the class-wise accuracy, F1-score, and AUC scores.

\begin{table}[t]
\centering
\caption{FID and MS-SSIM scores for benign and malignant cases with 30k samples.}
\label{tab:FID_results}
\begin{tabular}{@{}lclc ccc@{}}
\toprule
\multirow{2}{*}{Metric} & \phantom{a} & \multirow{2}{*}{Compared Dataset} & \phantom{a} & All & Benign & Malignant \\ 
&&&& (60,000) & (30,000) & (30,000) \\
\midrule
\multirow{2}{*}{FID} && ISIC + Fitzpatrick && 86.53 & 25.77  & 117.01          \\
&& Fitzpatrick && 20.34        & 79.14          & 15.35     \\ 
\midrule
MS-SSIM &  &-- & & 0.35 & 0.46 & 0.27 \\
\bottomrule
\end{tabular}%
\end{table}

\begin{table}[t]
\centering
\caption{Skin tone detection performance comparison on the DDI dataset when the model is trained on different datasets. Accuracy and F-1 scores are reported for each skin tone: A-lighter skin tone, B-Brown skin tone, and C-darker skin tone.}
\label{tab:skin_tone_results}
\setlength{\tabcolsep}{4pt}
\begin{tabular}{@{}lc ccc c ccc@{}}
\toprule
\multirow{2}{*}{Approach} & \phantom{a} & \multicolumn{3}{c}{Accuracy} & \phantom{a} & \multicolumn{3}{c}{F1-score} \\ 
\cmidrule{3-5}\cmidrule{7-9} 

&& A & B & C && A & B & C \\ 
\midrule
ITA  && \textbf{0.89} & 0.14  & 0.18 && 0.51 & 0.22 & 0.27\\
Real && 0.84 & \textbf{0.62} & 0.40 && \textbf{0.82} & \textbf{0.63} & 0.44 \\
Synthetic && 0.58 & 0.51 & \textbf{0.57} && 0.55 & 0.50 & 0.61\\
Real+Synthetic && 0.67 & 0.60 & 0.50 && 0.62 & 0.54 & \textbf{0.64}\\ 
\bottomrule
\end{tabular}%
\end{table}

\begin{table}[t]
\centering
\caption{Skin disease classification results on DDI dataset.}
\label{tab:ddi_skin_disease_results}
\setlength{\tabcolsep}{4pt}
\resizebox{\linewidth}{!}{%
\begin{tabular}{@{}lc ccc c ccc c ccc@{}}
\toprule
\multirow{2}{*}{Training data} & \phantom{a} & \multicolumn{3}{c}{Accuracy} & \phantom{a} & \multicolumn{3}{c}{F1-score} & \phantom{a} & \multicolumn{3}{c}{AUC} \\ 
\cmidrule{3-5}\cmidrule{7-9} \cmidrule{11-13}

&& A & B & C && A & B & C && A & B & C\\ 
\midrule
Fitzpatrick && 0.63 & 0.61 & 0.56  && 0.59 & 0.60 & 0.49 && 0.70 & 0.71 & 0.51\\
ISIC && \textbf{0.76} & 0.69 & \textbf{0.76} && 0.43 & 0.41 & 0.43 && 0.58 & 0.67 & 0.52\\
Real+Synthetic && 0.71 & \textbf{0.77} & 0.68 && \textbf{0.63} & \textbf{0.72} & \textbf{0.52} && \textbf{0.72} & \textbf{0.78} & \textbf{0.58}\\ 
\bottomrule
\end{tabular}%
}
\end{table}

\begin{table}[t]
\centering
\caption{Skin disease classification results on Atlas dataset, MClass dataset, and ISIC-2018 test dataset. We present accuracy, F1-score, and AUC across three skin tones.}
\label{tab:atlas_skin_disease_results}
\setlength{\tabcolsep}{4pt}
\resizebox{\linewidth}{!}{%
\begin{tabular}{@{}ll cc ccc c ccc c ccc@{}}
\toprule
\multirow{2}{*}{Test data} & \multirow{2}{*}{Training data} & \phantom{a} & \multicolumn{3}{c}{Accuracy} & \phantom{a} & \multicolumn{3}{c}{F1-score} & \phantom{a} & \multicolumn{3}{c}{AUC} \\ 
\cmidrule{4-6}\cmidrule{8-10} \cmidrule{12-14}

&& & A & B & C && A & B & C && A & B & C\\ 
\midrule
\multirow{3}{*}{Atlas Derm} & Fitzpatrick && 0.50 & 0.56 & 0.58  && 0.49 & 0.56 & 0.55 && 0.71 & 0.63 & 0.61\\
& ISIC && \textbf{0.78} & 0.60 & 0.42 && 0.45 & 0.41 & 0.36 && \textbf{0.77} & \textbf{0.75} & 0.63\\

& Real+Synthetic && 0.65 & \textbf{0.61} & \textbf{0.68} && \textbf{0.69} & \textbf{0.61} & \textbf{0.63} && 0.69 & 0.65 & \textbf{0.68}\\
\midrule
\multirow{3}{*}{Atlas Clinical} & Fitzpatrick && 0.43 & 0.42 & 0.52  && 0.43 & 0.41 & 0.50 && \textbf{0.68} & 0.64 & 0.58\\

& ISIC && \textbf{0.78} & \textbf{0.70} & 0.40 && \textbf{0.45} & 0.41 & 0.29 && 0.64 & 0.53 & 0.54\\

& Real+Synthetic && 0.41 & 0.67 & \textbf{0.64} && 0.41 & \textbf{0.65} & \textbf{0.60} && 0.61 & \textbf{0.69} & \textbf{0.70}\\ 
\midrule
\multirow{3}{*}{MClass Derm} & Fitzpatrick && 0.71 & 0.67 & --  && \textbf{0.63} & 0.67 & -- && 0.69 & 1.0 & --\\

& ISIC && \textbf{0.83} & 0.33 & -- && 0.55 & 0.25 & -- && \textbf{0.77} & 1.0 & --\\

& Real+Synthetic && 0.77 & \textbf{1.0} & -- && 0.59 & \textbf{1.0} & -- && 0.66 & 1.0 & --\\ 

\midrule
\multirow{3}{*}{MClass Clinical} & Fitzpatrick && 0.65 & 0.64 & --  && 0.61 & \textbf{0.63} & -- && 0.77 & 0.77 & --\\

& ISIC && \textbf{0.82} & \textbf{0.75} & -- && 0.45 & 0.43 & -- && 0.79 & \textbf{0.75} & --\\

& Real+Synthetic && 0.70 & 0.46 & -- && \textbf{0.63} & 0.46 & -- && \textbf{0.81} & 0.71 & --\\ 
\midrule
\multirow{3}{*}{ISIC-2018} & Fitzpatrick && 0.54 & 0.62 & 1.0  && 0.47 & 0.61 & 1.0 && 0.71 & 0.80 & 1.0\\
&ISIC && \textbf{0.89} & \textbf{0.84} & 0.67 && \textbf{0.49} & \textbf{0.70} & 0.40 && \textbf{0.81} & \textbf{0.97} & 1.0\\
&Real+Synthetic && 0.60 & 0.46 & 1.0 && 0.48 & 0.45 & 1.0 && 0.61 & 0.57 & 1.0\\ 

\bottomrule
\end{tabular}%
}
\end{table}

\begin{figure}
  \centering
  \begin{tabular}{cccc}
    \includegraphics[width=0.25\textwidth]{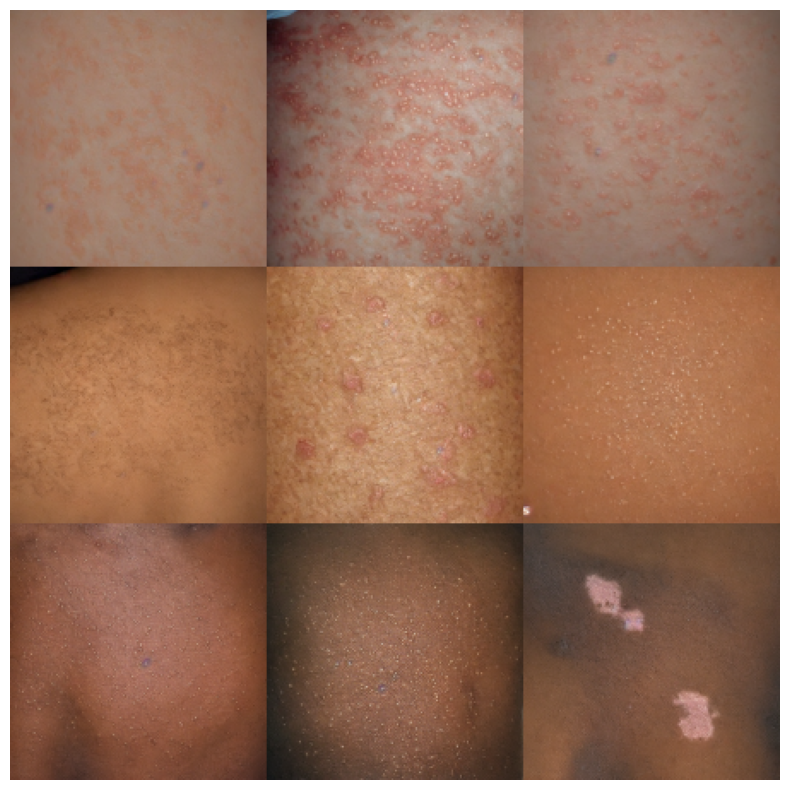} &
    \includegraphics[width=0.25\textwidth]{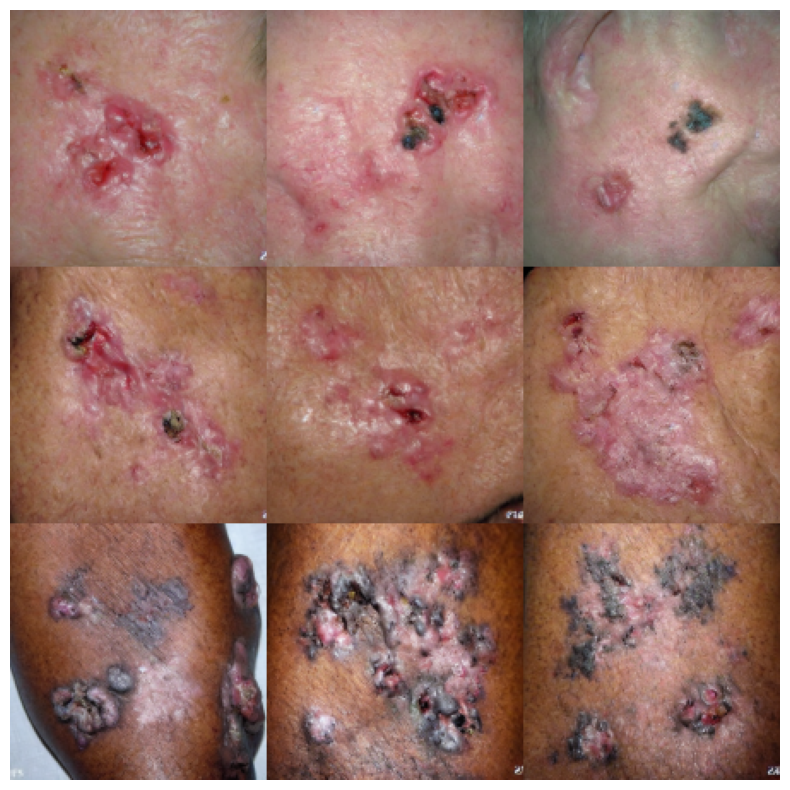} &
    \includegraphics[width=0.25\textwidth]{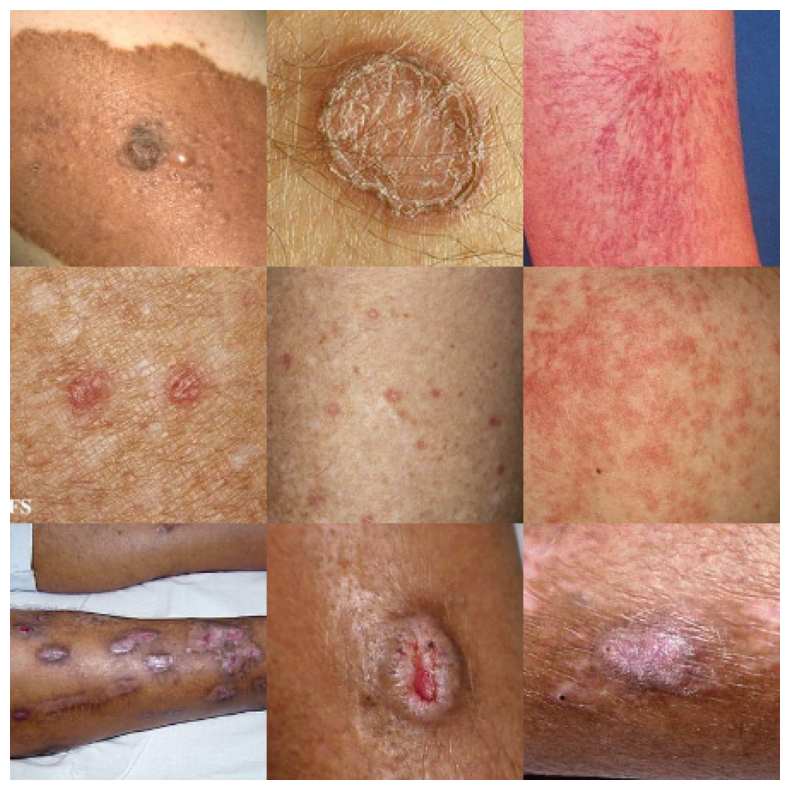} &
    \includegraphics[width=0.25\textwidth]{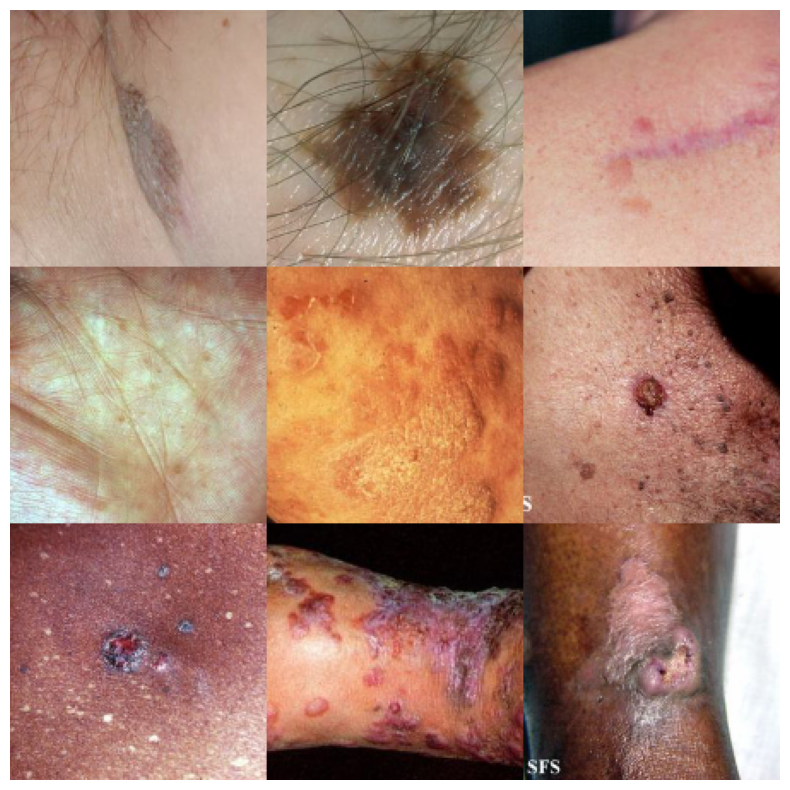} \\
  (a) & (b) & (c) & (d)\\
  \end{tabular}
  \caption{Visual comparison of the dermoscopic images in the Fitzpatrick17k dataset with the DermDiff-generated samples. Rows 1-3 denote skin tones A-C. Examples from DermDiff generated images and real images. DermDiff generated (a) benign and (b) malignant images; Fitzpatrick17k (c) benign and (d) malignant images.}
  \label{fig:sample_images}
\end{figure}

\subsection{Results and Discussion}
\textbf{Generation of Synthetic Dermoscopic Images:} 
We trained our Dermdiff model with the Fitzpatrick dataset and generated a total number of 60k synthetic images. Among them 30k were benign (10k for each skin tone) and similarly 30k malignant (10k for each skin tone) images. We evaluated our DermDiff model based on fidelity by calculating the FID score, which calculates the distance between real data distribution and synthetic data distribution. A lower FID score means generated images have good similarity with the distribution of original images. Although the FID score is calculated using an Inception V3 model, trained on the ImageNet dataset, we fine-tuned the Inception V3 model by training a disease classification model with the dermatology dataset so that it can provide a better representation of the dataset. We report individual FID scores for benign and malignant cases in Table~\ref{tab:FID_results}. Although we trained our generative model with only the Fitzpatrick dataset, we calculated the FID score comparing both Fitzpatrick and ISIC data. That is why we got a lower FID score when compared to the Fitzpatrick dataset and a comparatively higher FID score when compared with the combined Fitzpatrick and ISIC datasets. We also calculated MS-SSIM scores individually for both benign and malignant cases. A lower MS-SSIM score shows less structural similarity between generated images, which may be interpreted as more variation among those newly generated images. Although we fine-tuned the model using a smaller dataset, our generated images are quite comparable to the real dataset images. Again the training dataset was not balanced across all skin tones but DermDiff generated a balanced dataset where an equal number of samples of each class is present. Fig.~\ref{fig:sample_images} visualizes sample images of different skin tones and disease classes, generated by the DermDiff model and from the Fitzpatrick dataset.

\begin{figure}[t]
  \centering
  \begin{tabular}{ccc}
    \includegraphics[width=0.33\textwidth, trim={0cm 0cm 0cm 1.35cm}, clip]{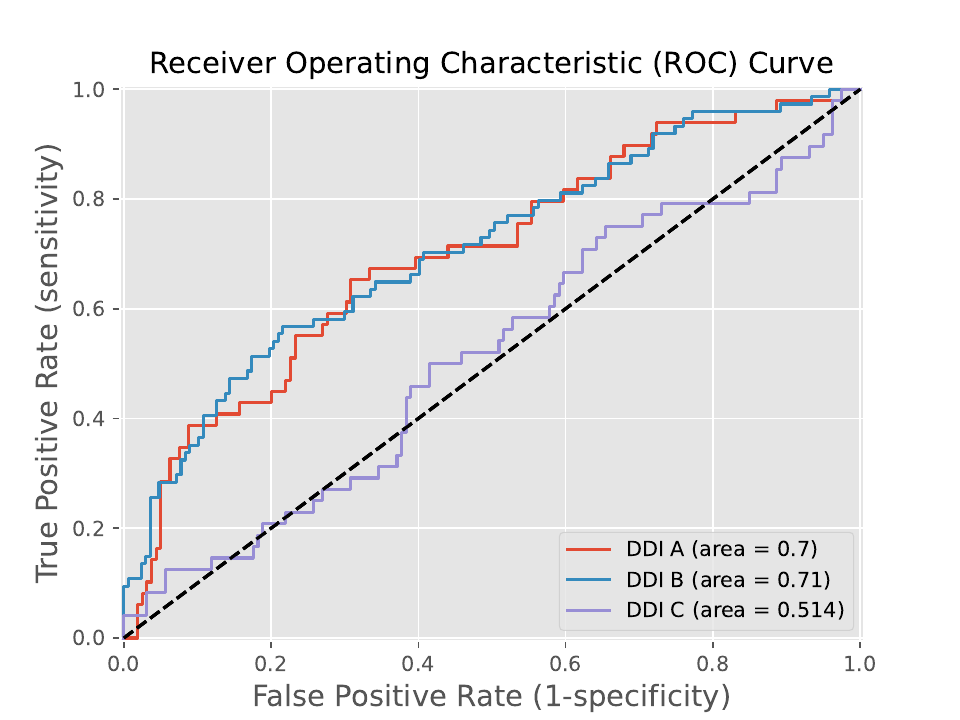} &
    \includegraphics[width=0.33\textwidth, trim={0cm 0cm 0cm 1.35cm}, clip]{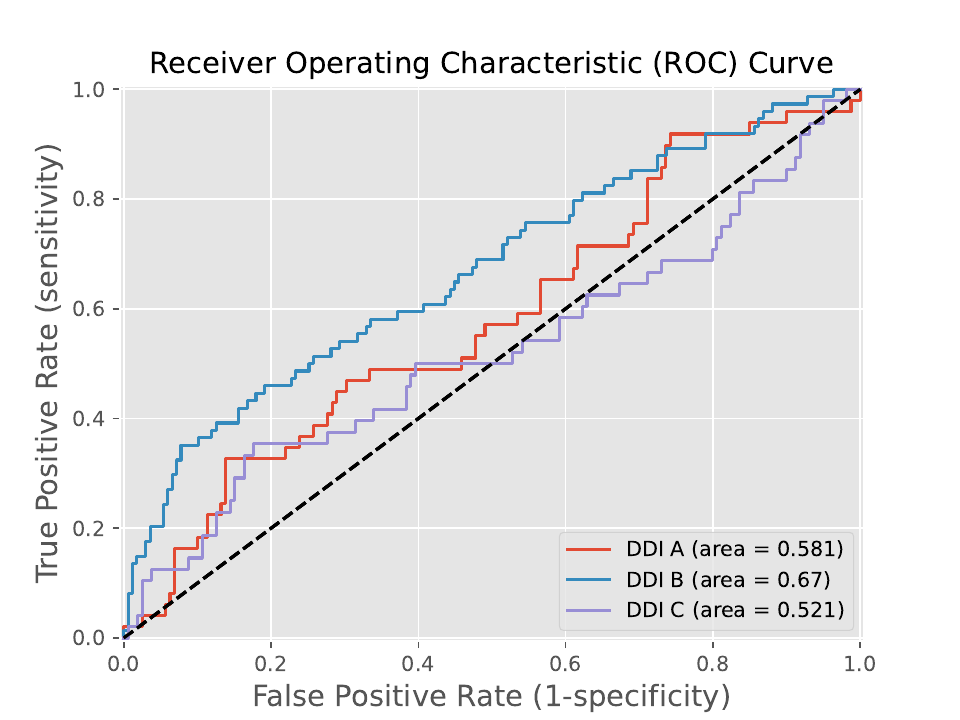} &
    \includegraphics[width=0.33\textwidth, trim={0cm 0cm 0cm 1.35cm}, clip]{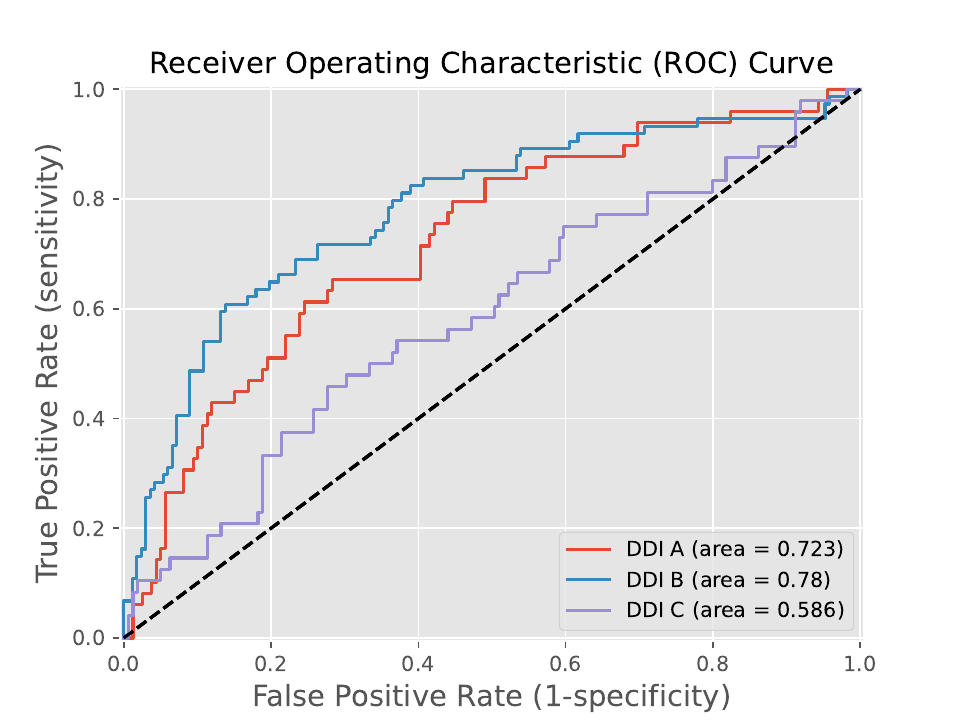} \\
  \end{tabular}
  \caption{Diagnostic performance on DDI dataset when the model was trained on (a) Fitzpatrick, (b) ISIC, and (c) combined real and synthetic image samples.}
  \label{fig:ROC_curve_DDI}
\end{figure}

\noindent\textbf{Detection of Skin Tones:}
For skin tone detection, we measured the accuracy score and F-1 score across all skin tones as reported in Table~\ref{tab:skin_tone_results}. We experimented with ITA-based skin tone prediction on the DDI dataset, which only performs well for lighter skin tones. We trained our skin tone detection model separately with with Fitzpatrick dataset, synthetic dataset, and combined both Fitzpatrick and synthetic datasets, and evaluated them on the DDI dataset. Our model trained with both Fitzpatrick and synthetic datasets performs better than others for darker skin tones. Although the performance for lighter skin tones is slightly degraded, it prioritizes darker skin tones. We also detect the skin tones for our test sets: ISIC-2018, Atlas, and MClass.

\noindent\textbf{Diagnosis of Skin Lesions:}
We trained our disease diagnosis model to classify benign and malignant cases from dermoscopic images. We trained our model separately with the Fitzpatrick dataset, and ISIC dataset and combined synthetic and real datasets. We evaluated our model by calculating Accuracy, F1-score, and AUC scores across all three skin tones A, B, and C. We predicted the skin tones for our test datasets so that we could evaluate our model across all three skin tones. The accuracy metric measures the percentage of correctly predicted cases, but from accuracy only we cannot predict how well the model is performing for benign and malignant cases individually. So, we measured the F1-score to get a balanced assessment of a model's performance by considering both precision and recall and the AUC score to evaluate the model's ability to distinguish between benign and malignant instances. We tested our disease diagnosis model on the DDI dataset, ISIC-2018 dataset, Atlas dataset, and MClass datasets. For Atlas and MClass datasets, we individually tested dermoscopic and clinical images. All the test results are reported in Tables~\ref{tab:ddi_skin_disease_results} and~\ref{tab:atlas_skin_disease_results}. For the DDI dataset, the trained model with the added synthetic data did not improve the accuracy but in terms of F1-score and AUC, it performed better than others across the skin tones (Fig.~\ref{fig:ROC_curve_DDI}). As synthetic data contained diverse class skin tone images, it improves performance for minority groups such as darker skin tones. The ISIC-2018 test set is a different than the Fitzpatrick dataset. Its performance was better only when trained with the ISIC train dataset, which has a similar distribution. For other datasets, this model performed better on darker skin tones and obtained better F1-scores than others. Following \citep{xu2022addressing}, we also measured several fairness metrics, including accuracy parity, equal opportunity, and equalized odds. Our results indicate that incorporating DermDiff-generated synthetic data improves these metrics for individuals with darker skin tones.

\section{Conclusion}
We have proposed a novel generative AI framework Dermdiff that can generate diverse dermoscopic images based on skin tones and disease statuses. Considering the effectiveness of DermDiff, it can facilitate the generation of diverse image datasets, particularly improving representations of underrepresented groups. We also found the newly generated images improve downstream diagnostic performance. Both skin tone detection and disease classification models performed better on marginalized groups when trained with combined real and DermDiff-generated synthetic data.

\section{Acknowledgement}
This work was funded by the UNITE Research Priority Area at the University of Kentucky.

\bibliographystyle{splncsnat}
\bibliography{references}

\begin{thebibliography}{25}
\providecommand{\natexlab}[1]{#1}
\providecommand{\url}[1]{\texttt{#1}}
\providecommand{\urlprefix}{}

\bibitem[{Almuzainit et~al.(2023)Almuzainit, Dendukuri, and Singh}]{safe}
Almuzainit, A.A., Dendukuri, S.K., Singh, V.K.: Toward fairness across skin tones in dermatological image processing.
\newblock In: 2023 IEEE 6th International Conference on (MIPR). pp. 1--7. IEEE (2023)

\bibitem[{Bevan and Atapour-Abarghouei(2022)}]{detecting_melanoma}
Bevan, P.J., Atapour-Abarghouei, A.: Detecting melanoma fairly: Skin tone detection and debiasing for skin lesion classification.
\newblock In: MICCAI Workshop on Domain Adaptation and Representation Transfer. pp. 1--11. Springer (2022)

\bibitem[{Brinker et~al.(2019)Brinker, Hekler, Hauschild, Berking, Schilling, Enk, Haferkamp, Karoglan, von Kalle, Weichenthal et~al.}]{mclass_data}
Brinker, T.J., Hekler, A., Hauschild, A., Berking, C., Schilling, B., Enk, A.H., Haferkamp, S., Karoglan, A., von Kalle, C., Weichenthal, M., et~al.: Comparing artificial intelligence algorithms to 157 german dermatologists: the melanoma classification benchmark.
\newblock European Journal of Cancer 111, 30--37 (2019)

\bibitem[{Chiu et~al.(2023)Chiu, Chung, Chen, Shi, and Ho}]{me_fairprune}
Chiu, C.H., Chung, H.W., Chen, Y.J., Shi, Y., Ho, T.Y.: Toward fairness through fair multi-exit framework for dermatological disease diagnosis.
\newblock arXiv preprint arXiv:2306.14518  (2023)

\bibitem[{Codella et~al.(2018)Codella, Gutman, Celebi, Helba, Marchetti, Dusza, Kalloo, Liopyris, Mishra, Kittler et~al.}]{isic2018skin}
Codella, N.C., Gutman, D., Celebi, M.E., Helba, B., Marchetti, M.A., Dusza, S.W., Kalloo, A., Liopyris, K., Mishra, N., Kittler, H., et~al.: Skin lesion analysis toward melanoma detection: A challenge at the 2017 international symposium on biomedical imaging (isbi), hosted by the international skin imaging collaboration (isic).
\newblock In: 2018 IEEE 15th international symposium on biomedical imaging (ISBI 2018). pp. 168--172. IEEE (2018)

\bibitem[{Daneshjou et~al.(2022)Daneshjou, Vodrahalli, Novoa, Jenkins, Liang, Rotemberg, Ko, Swetter, Bailey, Gevaert et~al.}]{ddi}
Daneshjou, R., Vodrahalli, K., Novoa, R.A., Jenkins, M., Liang, W., Rotemberg, V., Ko, J., Swetter, S.M., Bailey, E.E., Gevaert, O., et~al.: Disparities in dermatology ai performance on a diverse, curated clinical image set.
\newblock Science advances 8(31), eabq6147 (2022)

\bibitem[{Fitzpatrick(1988)}]{fitzpatrick1988validity}
Fitzpatrick, T.B.: The validity and practicality of sun-reactive skin types i through vi.
\newblock Archives of dermatology 124(6), 869--871 (1988)

\bibitem[{Groh et~al.(2021)Groh, Harris, Soenksen, Lau, Han, Kim, Koochek, and Badri}]{groh2021fitzpatrick}
Groh, M., Harris, C., Soenksen, L., Lau, F., Han, R., Kim, A., Koochek, A., Badri, O.: Evaluating deep neural networks trained on clinical images in dermatology with the fitzpatrick 17k dataset.
\newblock In: Proceedings of the IEEE/CVF conference on CVPR. pp. 1820--1828 (2021)

\bibitem[{Imran and Terzopoulos(2021)}]{imran2021multi}
Imran, A.A.Z., Terzopoulos, D.: Multi-adversarial variational autoencoder nets for simultaneous image generation and classification.
\newblock Deep Learning Applications, Volume 2 pp. 249--271 (2021)

\bibitem[{Kawahara et~al.(2018)Kawahara, Daneshvar, Argenziano, and Hamarneh}]{atlas2018seven}
Kawahara, J., Daneshvar, S., Argenziano, G., Hamarneh, G.: Seven-point checklist and skin lesion classification using multitask multimodal neural nets.
\newblock IEEE journal of biomedical and health informatics 23(2), 538--546 (2018)

\bibitem[{Kingma and Welling(2013)}]{vae_enc}
Kingma, D.P., Welling, M.: Auto-encoding variational bayes.
\newblock arXiv preprint arXiv:1312.6114  (2013)

\bibitem[{Kinyanjui et~al.(2019)Kinyanjui, Odonga, Cintas, Codella, Panda, Sattigeri, and Varshney}]{ita_skin_tone}
Kinyanjui, N.M., Odonga, T., Cintas, C., Codella, N.C., Panda, R., Sattigeri, P., Varshney, K.R.: Estimating skin tone and effects on classification performance in dermatology datasets.
\newblock preprint arXiv:1910.13268  (2019)

\bibitem[{Kinyanjui et~al.(2020)Kinyanjui, Odonga, Cintas, Codella, Panda, Sattigeri, and Varshney}]{kinyanjui2020fairness}
Kinyanjui, N.M., Odonga, T., Cintas, C., Codella, N.C., Panda, R., Sattigeri, P., Varshney, K.R.: Fairness of classifiers across skin tones in dermatology.
\newblock In: International Conference on MICCAI. pp. 320--329. Springer (2020)

\bibitem[{Miko{\l}ajczyk et~al.(2022)Miko{\l}ajczyk, Majchrowska, and Carrasco~Limeros}]{de_biasing2022}
Miko{\l}ajczyk, A., Majchrowska, S., Carrasco~Limeros, S.: The (de) biasing effect of gan-based augmentation methods on skin lesion images.
\newblock In: International Conference on Medical Image Computing and Computer-Assisted Intervention. pp. 437--447. Springer (2022)

\bibitem[{Ramesh et~al.(2022)Ramesh, Dhariwal, Nichol, Chu, and Chen}]{clip_text}
Ramesh, A., Dhariwal, P., Nichol, A., Chu, C., Chen, M.: Hierarchical text-conditional image generation with clip latents.
\newblock arXiv preprint arXiv:2204.06125 1(2), 3 (2022)

\bibitem[{Rombach et~al.(2022{\natexlab{a}})Rombach, Blattmann, Lorenz, Esser, and Ommer}]{rombach2022high}
Rombach, R., Blattmann, A., Lorenz, D., Esser, P., Ommer, B.: High-resolution image synthesis with latent diffusion models.
\newblock In: Proceedings of the IEEE/CVF Conference on Computer Vision and Pattern Recognition. pp. 10684--10695 (2022{\natexlab{a}})

\bibitem[{Rombach et~al.(2022{\natexlab{b}})Rombach, Blattmann, Lorenz, Esser, and Ommer}]{stable_diff}
Rombach, R., Blattmann, A., Lorenz, D., Esser, P., Ommer, B.: High-resolution image synthesis with latent diffusion models.
\newblock In: Proceedings of the IEEE/CVF Conference on CVPR. pp. 10684--10695 (2022{\natexlab{b}})

\bibitem[{Rotemberg et~al.(2021)Rotemberg, Kurtansky, Betz-Stablein, Caffery, Chousakos, Codella, Combalia, Dusza, Guitera, Gutman et~al.}]{isic2020}
Rotemberg, V., Kurtansky, N., Betz-Stablein, B., Caffery, L., Chousakos, E., Codella, N., Combalia, M., Dusza, S., Guitera, P., Gutman, D., et~al.: A patient-centric dataset of images and metadata for identifying melanomas using clinical context. (2021)

\bibitem[{Schuhmann et~al.(2021)Schuhmann, Vencu, Beaumont, Kaczmarczyk, Mullis, Katta, Coombes, Jitsev, and Komatsuzaki}]{laion}
Schuhmann, C., Vencu, R., Beaumont, R., Kaczmarczyk, R., Mullis, C., Katta, A., Coombes, T., Jitsev, J., Komatsuzaki, A.: Laion-400m: Open dataset of clip-filtered 400 million image-text pairs.
\newblock arXiv preprint arXiv:2111.02114  (2021)

\bibitem[{Society(2024)}]{ACS2024}
Society, A.C.: Cancer Facts \& Figures 2024.
\newblock American Cancer Society, Atlanta (2024)

\bibitem[{Tschandl et~al.(2018)Tschandl, Rosendahl, and Kittler}]{Tschandl2018_HAM10000}
Tschandl, P., Rosendahl, C., Kittler, H.: The {HAM10000} dataset, a large collection of multi-source dermatoscopic images of common pigmented skin lesions.
\newblock Sci. Data 5, 180161 (2018)

\bibitem[{Wu et~al.(2022)Wu, Zeng, Xu, Shi, and Hu}]{fairprune}
Wu, Y., Zeng, D., Xu, X., Shi, Y., Hu, J.: Fairprune: Achieving fairness through pruning for dermatological disease diagnosis.
\newblock In: International Conference on MICCAI. pp. 743--753. Springer (2022)

\bibitem[{Xie et~al.(2017)Xie, Girshick, Doll{\'a}r, Tu, and He}]{resnext}
Xie, S., Girshick, R., Doll{\'a}r, P., Tu, Z., He, K.: Aggregated residual transformations for deep neural networks.
\newblock In: Proceedings of the IEEE conference on computer vision and pattern recognition. pp. 1492--1500 (2017)

\bibitem[{Xu et~al.(2022)Xu, Li, Yao, Li, Zhao, and Zhou}]{xu2022addressing}
Xu, Z., Li, J., Yao, Q., Li, H., Zhao, M., Zhou, S.K.: Addressing fairness issues in deep learning-based medical image analysis: A systematic review (2022)

\bibitem[{Xu et~al.(2023)Xu, Zhao, Quan, Yao, and Zhou}]{fairadabn}
Xu, Z., Zhao, S., Quan, Q., Yao, Q., Zhou, S.K.: Fairadabn: Mitigating unfairness with adaptive batch normalization and its application to dermatological disease classification.
\newblock arXiv preprint arXiv:2303.08325  (2023)

\end{thebibliography}

\end{document}